\documentclass{article}

\usepackage{microtype}
\usepackage{graphicx}
\usepackage{stackengine}
\usepackage{booktabs} 

\usepackage{hyperref}

\usepackage[accepted]{icml2023}

\usepackage{amsmath}
\usepackage{amssymb}
\usepackage{mathtools}
\usepackage{amsthm}
\usepackage{bm}

\usepackage[capitalize,noabbrev]{cleveref}

\theoremstyle{plain}

\theoremstyle{definition}

\theoremstyle{remark}

\usepackage{chngcntr}
\counterwithin{table}{section}

\usepackage[textsize=tiny]{todonotes}

\icmltitlerunning{Exponential weight averaging as damped harmonic motion}

\newcommand{\methodname}{\texttt{BELAY}}

\begin{document}

\twocolumn[
\icmltitle{Exponential weight averaging as damped harmonic motion}



\icmlsetsymbol{equal}{*}

\begin{icmlauthorlist}
\icmlauthor{Jonathan Patsenker}{equal,yale}
\icmlauthor{Henry Li}{equal,yale}
\icmlauthor{Yuval Kluger}{yale}
\end{icmlauthorlist}

\icmlaffiliation{yale}{Applied Math Program, Yale University, New Haven, CT, USA}

\icmlcorrespondingauthor{Henry Li}{henry.li@yale.edu}
\icmlcorrespondingauthor{Jonathan Patsenker}{jonathan.patsenker@yale.edu}

\icmlkeywords{Machine Learning, ICML}

\vskip 0.3in
]

\begin{abstract}
    The exponential moving average (EMA) is a commonly used statistic for providing stable estimates of stochastic quantities in deep learning optimization. Recently, EMA has seen considerable use in generative models, where it is computed with respect to the model weights, and significantly improves the stability of the inference model during and after training. While the practice of weight averaging at the end of training is well-studied and known to  improve estimates of local optima, the benefits of EMA over the course of training is less understood. In this paper, we derive an explicit connection between EMA and a damped harmonic system between two particles, where one particle (the EMA weights) is drawn to the other (the model weights) via an idealized zero-length spring. We then leverage this physical analogy to analyze the effectiveness of EMA, and propose an improved training algorithm, which we call \methodname{}. Finally, we demonstrate theoretically and empirically several advantages enjoyed by \methodname{} over standard EMA.
\end{abstract}

\section{Introduction}

\begin{figure}
    \centering
    \includegraphics[width=0.9\linewidth]{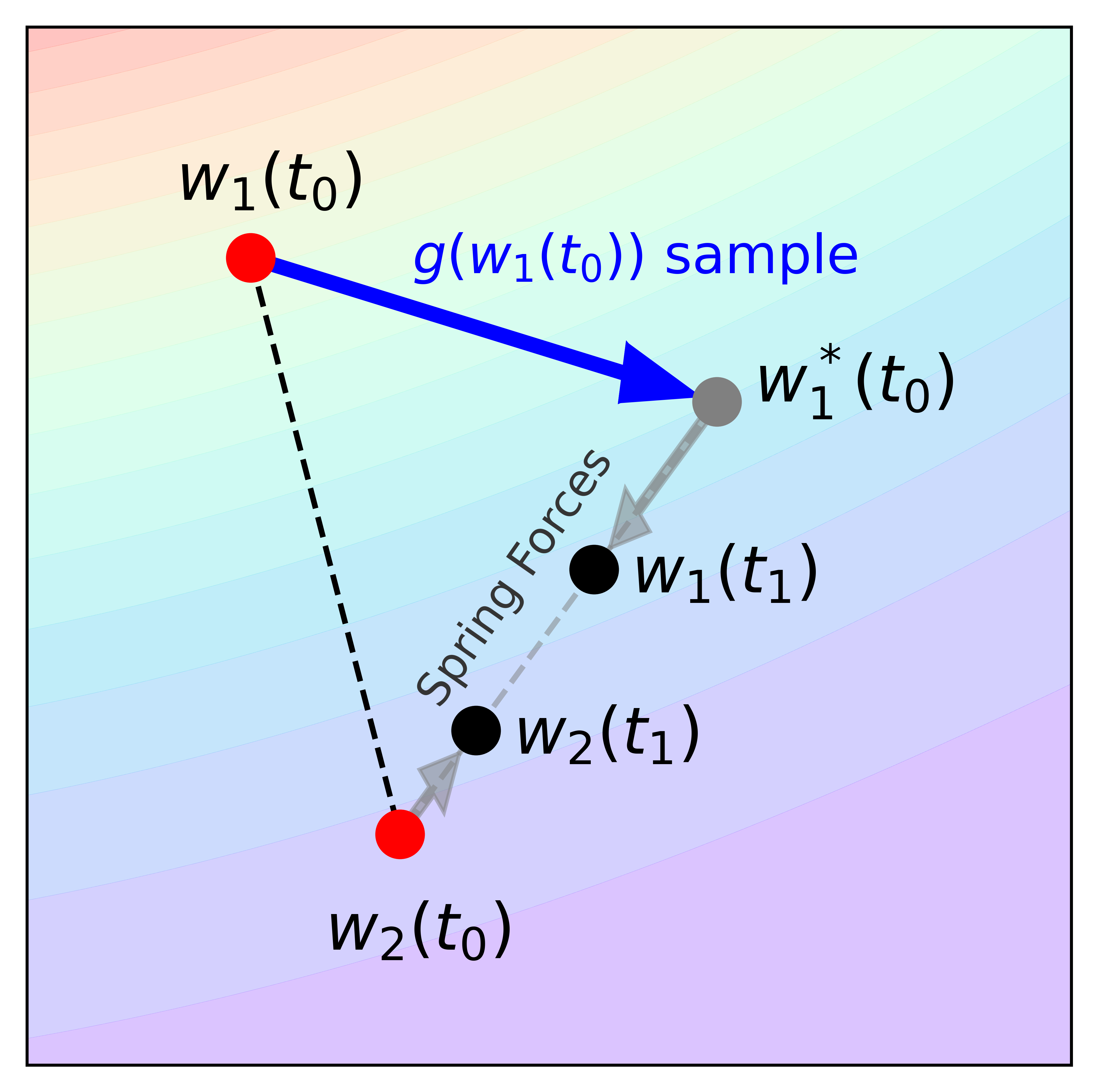}
    \caption{A visualization of the \methodname{} update step. The background color corresponds to the true full-batch loss function, and $g$ is sampled using an optimizer on a minibatch.}
    \label{fig:belay}
\end{figure}

First-order stochastic gradient optimizers are widely employed in the modern deep learning literature to train large models, and are often motivated by physical analogies such as momentum \cite{polyak1964some}, curvature \cite{ypma1995historical}, projections \cite{hestenes1952methods,beck2003mirror}, and molecular dynamics \cite{bussi2007accurate,welling2011bayesian}.


We turn our attention to another common motif in first-order optimization methods, model stabilization via the exponential moving average (EMA). Applied to the gradient of the loss function, EMA is well-understood as a physical approximation of momentum, and this concept is extensively used in many popular adaptive algorithms \cite{kingma2014adam,hinton2012neural,zeiler2012adadelta,zhuang2020adabelief} to stabilize gradient descent. We instead focus on EMA applied directly to the model weights. This type of averaging is frequently used in deep learning, especially in generative modeling \cite{yazici2018unusual,ho2019flow++,song2020improved}. Weight-based EMA is related to the general idea of weight averaging \cite{ruppert1988efficient,polyak1992acceleration}, which is known to improve the generalization properties of stochastic algorithms at the end of training \cite{izmailov2018averaging}. However, a physical analogy for weight averaging has, to our knowledge, not been explored.

In this paper, we establish a clear connection between the weight-based EMA update and the discrete time Euler update of a damped harmonic oscillator. In other words, EMA can be modeled by an idealized zero-length spring that is attached on one end to the model weights, and on the other to the EMA weights. This analogy allows us to highlight several distinct theoretical properties of the weight-based EMA. Finally, we propose \methodname{}, which explores a variant of EMA where the model weights can also be updated by the EMA weights, and find that \methodname{} confers increased robustness of the optimization algorithm to the learning rate.



\section{Background}

For a deep neural network with parameters \(\bm{w} \in \mathbb{R}^n \), we aim to minimize some loss function $\mathcal{L}: \bm{w} \mapsto \mathbb{R}$. For applications of generative modeling, $\mathcal{L}$ is often negative log-likelihood \cite{salimans2017pixelcnn++,dinh2014nice} or a score matching loss \cite{hyvarinen2005estimation,vincent2011connection,song2020sliced,ho2020denoising}. First-order stochastic gradient optimizers iteratively sample an estimate for $\nabla \mathcal{L}(\bm{w})$ and update $\bm{w}$ with this information. In this work, we refer to this 'update' for an optimizer as a function $g(\bm{w}) \in \mathbb{R}^n$.


\subsection{The Exponential Moving Average}

Using the time-averaged parameters of a model over the course of training is a commonly used technique in generative modeling across GANs \cite{salimans2016improved,yazici2018unusual}, normalizing flows \cite{ho2019flow++}, autoregressive models \cite{salimans2017pixelcnn++,child2019generating}, and diffusion models \cite{song2020improved,song2020score,ho2020denoising}. The exponential moving average (EMA) of a set of parameters $\bm{w}$ over time can be described recursively in terms of the time update
\begin{align}
    \begin{split}
        \label{eq:ema}
        \bm{w}^{EMA}(t+1) &= \alpha \bm{w}^*(t) + (1 - \alpha) \bm{w}^{EMA}(t)\\
        \bm{w}^*(t) &= \bm{w}(t) + g(\bm{w}(t))
    \end{split}
\end{align}
where $\alpha \in [0, 1]$. Weighted-based EMA has been shown to improve model stability over the course of training \cite{song2020improved,yazici2018unusual}.


\subsection{Damped Harmonic Oscillators}

Let $\bm{w}_1, \bm{w}_2: [0,T] \rightarrow \mathbb{R}^n$ denote the positions of two point particles with masses $m_1, m_2$, connected by an idealized zero-length spring. In a damped harmonic system, the force exerted by $w_2$ on $w_1$ can be written as
\begin{equation}
    \bm{F} = \underbrace{k(\bm{w}_2 - \bm{w}_1)}_{\bm{F}_S} - \underbrace{c_1\dot{\bm{w}}_1}_{\bm{F}_D},
\end{equation}
where $\bm{F}_S$ is the spring force given by Hooke's Law and spring constant $k$, and $\bm{F}_D$ is the damping force. External forces on $\bm{w}_1$ may be modeled with a function $f: \mathbb{R}^n \times [0,T] \rightarrow \mathbb{R}^n$. Applying Newton's second law of motion provides the ODE
\begin{equation}
    \ddot{\bm{w}}_1 = \frac{k}{m_1}(\bm{w}_2 - \bm{w}_1) - \frac{c_1}{m_1} \dot{\bm{w}}_1 + \frac{1}{m_1} f(\bm{w}_1, t).
\end{equation}
Using the second kinematics equation and discretizing $t$ with time-steps of size $\Delta t$, we have
\begin{align}
    \begin{split}
    \label{eq:kinematics}
        \bm{w}_1(t + \Delta t) &= \bm{w}_1(t) + \dot{\bm{w}}_1(t) \Delta t + \frac{1}{2}\ddot{\bm{w}}_1(t)\Delta t^2\\
        &= \bm{w}_1(t) + \frac{\Delta t^2}{2m_1} f(\bm{w}_1(t), t) \\
        &\quad + \frac{k \Delta t^2}{2 m_1}(\bm{w}_2(t) - \bm{w}_1(t)) \\
        &\quad + \left ( \Delta t - \frac{c_1 \Delta t^2}{2 m_1} \right )\dot{\bm{w}}_1(t)
    \end{split}
\end{align}
The position of the other mass, $\bm{w}_2$ can be similarly derived. For the purposes of this work, $\bm{w}_2$ is not subject to any external force. We thus have 
\begin{align}
    \begin{split}
    \label{eq:kinematics2}
        \bm{w}_2(t + \Delta t) &= \bm{w}_2(t) + \frac{k \Delta t^2}{2 m_2}(\bm{w}_1(t) - \bm{w}_2(t)) \\
        &\quad + \left ( \Delta t - \frac{c_2 \Delta t^2}{2 m_2} \right )\dot{\bm{w}}_2(t),
    \end{split}
\end{align}
where $c_2 \in \mathbb{R}^+$ is a damping constant analogous to $c_1$. Together these two equations can be used to describe to perform Euler integration and solve an initial value problem when $\bm{w}_1(0), \bm{w}_2(0)$ are known.

\section{EMA as Damped Harmonic Motion}
\label{sec:belay}

We relate Eqs. \ref{eq:kinematics} and \ref{eq:kinematics2} to weight-based EMA, described in Eq. \ref{eq:ema}, by choosing the damping parameter $c_1 = \frac{2m_1}{\Delta t}$, and setting $\beta = 1 - \frac{k \Delta t^2}{2 m_1}$. Thus, we can rewrite Equation \ref{eq:kinematics} as
\begin{align}
    \begin{split}
    \label{eq:k_ema_1}
        \bm{w}_1(t + \Delta t) &= \bm{w}_1(t) + \beta g'(\bm{w}_1(t)) + (1-\beta) (\bm{w}_2 (t)- \bm{w}_1 (t))\\
        &= \beta \bm{w}_1(t) + \beta g'(\bm{w}_1(t)) + (1-\beta) \bm{w}_2 (t)\\
        &= \beta \bm{w}^*_1(t) + (1-\beta) \bm{w}_2 (t),\\
    \end{split}
\end{align}
where $g' = \frac{\Delta t^2}{2\beta m_1}f(\cdot ,t)$, and $\bm{w}^*_1(t) = \bm{w}_1(t) + g'(\bm{w}_1(t))$.

Selecting $c_2 = \frac{2m_2}{\Delta t}$, and $\alpha = 1 - \frac{k \Delta t^2}{2 m_2}$, we can also rewrite Equation \ref{eq:kinematics2} as
\begin{equation}
    \bm{w}_2(t + \Delta t) = \alpha \bm{w}_1(t) + (1-\alpha) \bm{w}_2 (t).
\end{equation}

Taking the point-masses $\bm{w}_1$, $\bm{w}_2$ to be the model weights $\bm{w}$ and $\bm{w}^{EMA}$ respectively, we obtain a weight averaging method that is precisely EMA when $\beta = 1$. This occurs when $m_1 \rightarrow \infty$. In this case, $\beta \rightarrow 1$ and Eq. \ref{eq:k_ema_1} becomes $\bm{w}_1 (t + \Delta t) = \bm{w}^*$, which follows line 2 of Eq. \ref{eq:ema} (and therefore Eq. \ref{eq:kinematics2} follows line 1 of Eq. \ref{eq:ema}). This implies that EMA has a physical interpretation as a damped harmonic oscillator.

According to our physical interpretation of the standard EMA scheme, $\bm{w}^*$ is not affected by $\bm{w}^{EMA}$ during training since it has infinite mass. In the next section, we explore the possibility of finite $m_1$. This allows us to harness the strong model stability properties of EMA to guide parameter updates during training, which we evaluate in Section \ref{sec:experiments}.

\paragraph{\methodname{}} We introduce \methodname{}: (B)ridging (E)xponentia(L) moving (A)verages with spring s(Y)stems, a novel class of optimization methods that generalizes weight averaging by re-expressing each step as a forward Euler integration update to a damped harmonic oscillator. \methodname{} is parameterized by $(k, m_1, m_2, c_1, c_2)$, and weight update function $g(\bm{w},t)$ provided by some existing optimizer. We add the parameter $\eta \in \mathbb{R}^+$ to designate learning rate for the optimizer. The physical interpretation of each parameter is described in Table \ref{param-table}. We formalize the \methodname{} algorithm in Algorithm \ref{alg:belay}, and visualize one update step in Figure \ref{fig:belay}. We note that when $c_1, c_2$, are set as described above, the implementation can be simplified to avoid storing or computing momentum terms $\bm{\dot{w}}(t)$. Even with a non-trivial set of damping parameters, the time and memory overhead of \methodname{} is negligible compared to conventional training, as is the case in \cite{izmailov2018averaging}.

\begin{algorithm}[h]
\caption{\methodname{}}
\label{alg:belay}
\begin{algorithmic}
\STATE \textbf{Input:} Parameters ($k$, $m_1$, $m_2$, $c_1$, $c_2$), weight update function $g$, learning rate $\eta$.
\STATE \textbf{Initialize} $t=0$
\STATE Initialize $\bm{w}_1(0)$, $\bm{w}_2(0)$ with small random values
\STATE Initialize $\bm{\dot{w}}_1(0) = 0$, $\bm{\dot{w}}_2(0) = 0$
\WHILE{stopping criterion not met}
    \STATE Compute weight-update $g(\bm{w}_1(t), t)$ \COMMENT{see Table \ref{param-table} for details}
    \STATE Compute optimizer step $\bm{w^*}_1 = \bm{w}_1 + \eta g(\bm{w}_1(t), t)$
    \STATE Compute momentum $\bm{M}_1 = (1-\frac{c_1}{2m_1}) \bm{\dot{w}}_1(t)$
    \STATE Compute momentum $\bm{M}_2 = (1-\frac{c_2}{2m_2}) \bm{\dot{w}}_2(t)$
    \STATE $\alpha = 1-\frac{k}{2m_1}$
    \STATE $\beta  = 1-\frac{k}{2m_2}$
    \STATE Update $\bm{w}_1(t+1) = \alpha\bm{w^*}_1(t) + (1-\alpha)\bm{w}_2(t) + \bm{M}_1$
    \STATE Update $\bm{w}_2(t+1) = \beta\bm{w}_2(t) + (1-\beta)\bm{w}_1(t) + \bm{M}_2$
    \STATE $\bm{\delta v}_{1} = \frac{k}{m_1} (\bm{w}_2 - \bm{w}_1) - \frac{c_1}{m_1}\bm{\dot{w}}_1(t) + \frac{\eta}{2\alpha}g(\bm{w}_1(t), t)$ \footnotemark
    \STATE $\bm{\delta v}_{2} = \frac{k}{m_2} (\bm{w}_1 - \bm{w}_2) - \frac{c_2}{m_2}\bm{\dot{w}}_2(t)$
    \STATE Update $\bm{\dot{w}}_1(t+1) = \bm{\dot{w}}_1 + \bm{\delta v}_{1}$
    \STATE Update $\bm{\dot{w}}_2(t+1) = \bm{\dot{w}}_2 + \bm{\delta v}_{2}$
    \STATE $t = t + 1$
\ENDWHILE
\end{algorithmic}
\end{algorithm}

\footnotetext{We scale $g$ by $\frac{\eta}{2 \alpha}$ because we earlier re-expressed $\frac{1}{2 \alpha m_1} g $ as $\eta g$ to simplify the update to $\bm{w}_1(t)$ (since $\Delta t = 1$). Since $\bm{\delta v}_1 = \bm{\ddot{w}}_1(t) = \frac{k}{m_1} (\bm{w}_2 - \bm{w}_1) - \frac{c_1}{m_1} \bm{\dot{w}}_1(t) + \frac{1}{m_1} g(\bm{w}_1(t), t)$, and we would like to retain scaling by $\eta$ as a parameter, we can use $\frac{\eta}{2 \alpha}$.}

\subsection{Connection to Optimizers with Momentum}
In \methodname{}, the damping coefficients $c_1$ and $c_2$ explicitly control the momentum term in every weight update step. Even when these coefficients are set to cancel the momentum (as described in Section \ref{sec:belay}) however, \methodname{} still uses momentum information. When running SGD with momentum, the weights are updated with the step,
\begin{align}
    \begin{split}
        \bm{v}(t) &= \lambda \nabla \mathcal{L}(\bm{w}(t)) + (1-\lambda) \bm{v}(t-1)\\
        \bm{w}(t+1) &= \bm{w}(t) + \alpha \bm{v}(t)\\
        &= \bm{w}(t) + \alpha \sum_{s=0}^t(1-\lambda)^s \lambda \nabla \mathcal{L}(\bm{w}(t-s))
    \end{split}
\end{align}
where $\mathcal{L}$ is the loss function being optimized and $\alpha \in [0,1]$. In the case where $\mathcal{L}$ is linear,
\begin{align}
    \begin{split}
        \sum_{s=0}^t a_s \nabla \mathcal{L}(\bm{w}(t-s)) &= \nabla \mathcal{L}\left ( \sum_{s=0}^t a_s \bm{w}(t-s) \right )\\
        &= \nabla \mathcal{L}\left ( \bm{w}^{EMA}(t) \right ),
    \end{split}
\end{align}
where $a_s = (1-\lambda)^s \lambda$. Given specific parameter choices, this is an update for \methodname{}. We have shown that the linearization of a \methodname{} update $\nabla \mathcal{L}(\bm{w}^{EMA}(t))$ is therefore equivalent to the momentum update term $\bm{v}(t)$. This relationship is analogous to that of SWA and FGE documented by \cite{izmailov2018averaging}.

\subsection{Deriving the Spring Constant $k$}
One caveat of vanilla EMA is its sensitivity to $\alpha$, which governs the exponential decay of the moving average. Picking $\alpha$ too large causes the average to converge too slowly, which can drastically extend the training time of a learning algorithm. Conversely, too small $\alpha$ will overly favor the present iterate, which at best increases model variance at evaluation-time, and at worst defeats the intended purpose of the moving average. The challenge of choosing a proper $\alpha$ is further complicated by its dependence on the total runtime of the training protocol, in terms of the number of gradient steps $T$.  

We leverage our physical analogy to harmonic oscillators to choose $k$ such that the system is invariant to $T$. We achieve this by examining the closed form solution of an overdamped spring system:
\begin{equation}
    \bm{w}(t) = C_1 e^{\left(-\delta + \sqrt{\delta^2 - \frac{k}{m}}\right) t} + C_2 e^{\left(-\delta - \sqrt{\delta^2 - \frac{k}{m}}\right) t} 
    \label{eq:overdamped_system}
\end{equation}
where $\delta = \frac{1}{\Delta t}$. Using the initial conditions $\dot{\bm{w}}(0) = 0$ and $\bm{w}(0) = \mathbf{x}_0$, we may obtain integration constants $C_1$ and $C_2$, and then solving for the function $k(T)$ such that $\bm{w}(T) = \bm{w}(T_0)$, we obtain $k \approx k_0\frac{T_0}{T}$. In our experiments, we let $k_0 = 1$ and $T_0 = 1e6$. From this, we obtain our proposed $T$-invariant spring constant $k(T) = \frac{C}{T}$, where $C = 1e6$.


\section{Numerical Experiments}
\label{sec:experiments}
In this section, we evaluate \methodname{} on a set of synthetic and real optimization tasks, and demonstrate several empirical properties of the convergence behavior of \methodname{}.

\begin{figure}[h!]
    \footnotesize
    \centering
    \begin{tabular}{c}
    \includegraphics[width=5cm, height=5cm, trim={.5cm 0 0 0}, clip]{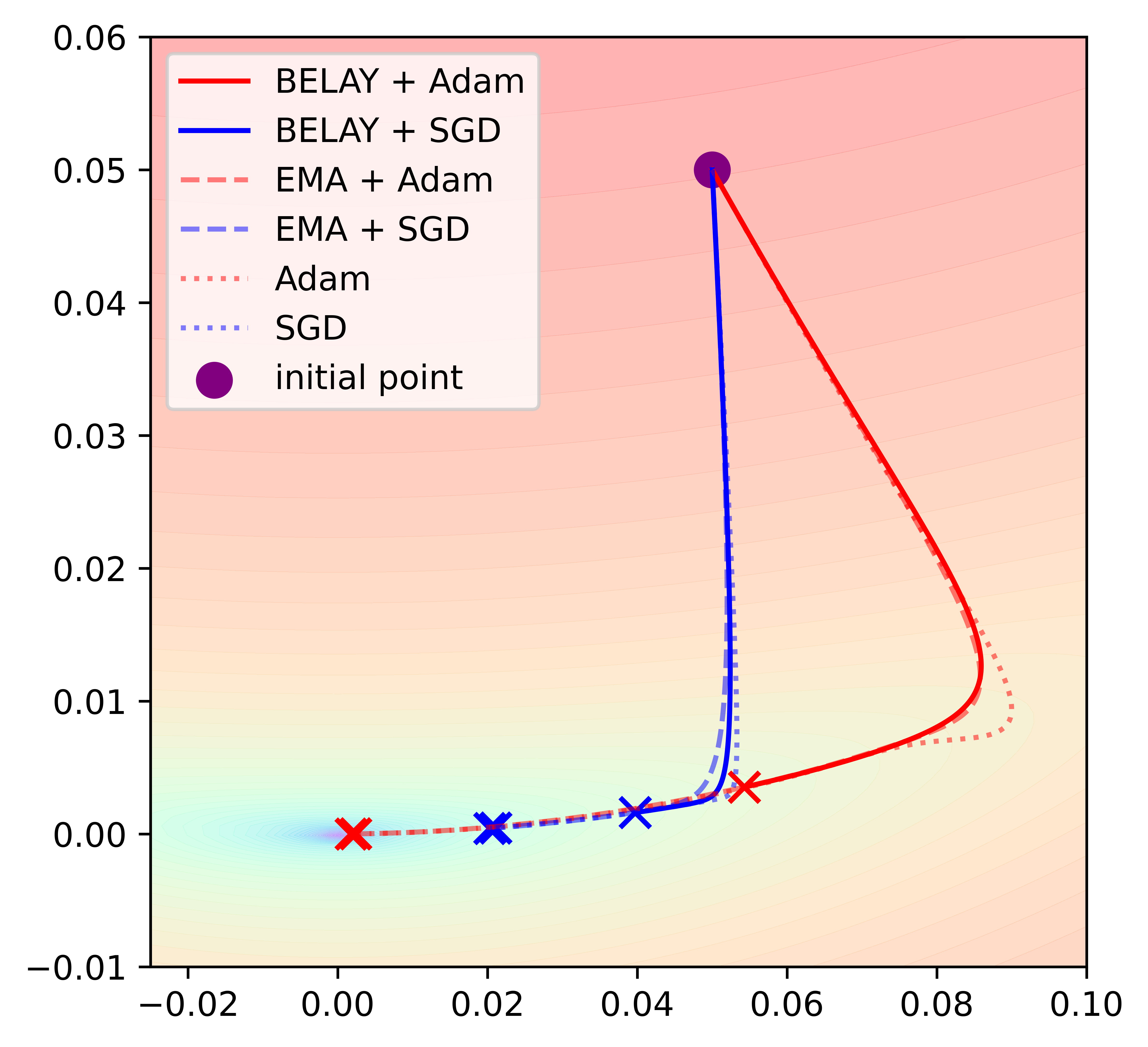} \\
    {\small lr = $0.001$} \\
    \includegraphics[width=5cm, height=5cm, trim={.5cm 0 0 0}, clip]{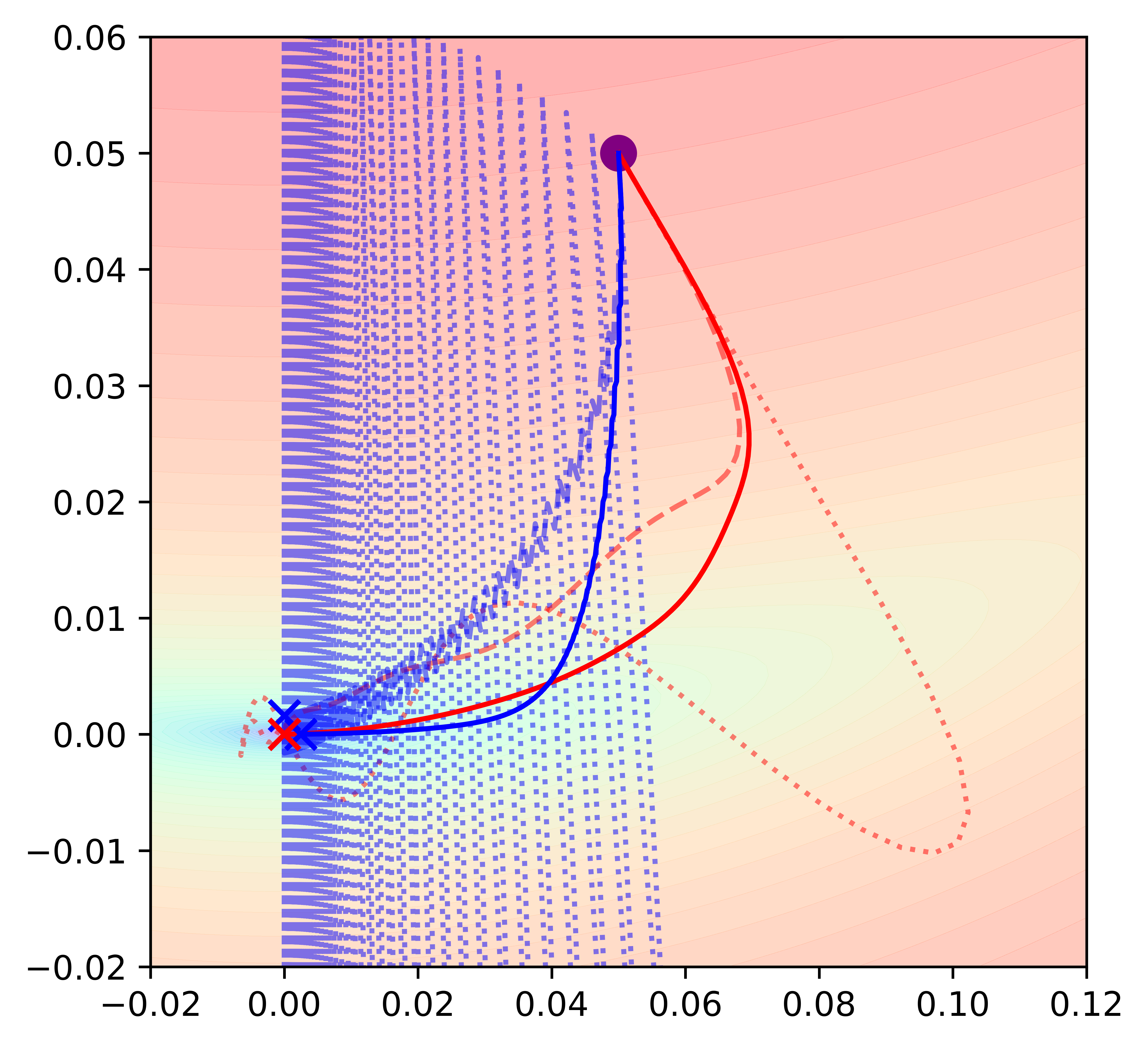} \\
    {\small lr = $0.01$} \\
    \includegraphics[width=5cm, height=5cm, trim={.5cm 0 0 0}, clip]{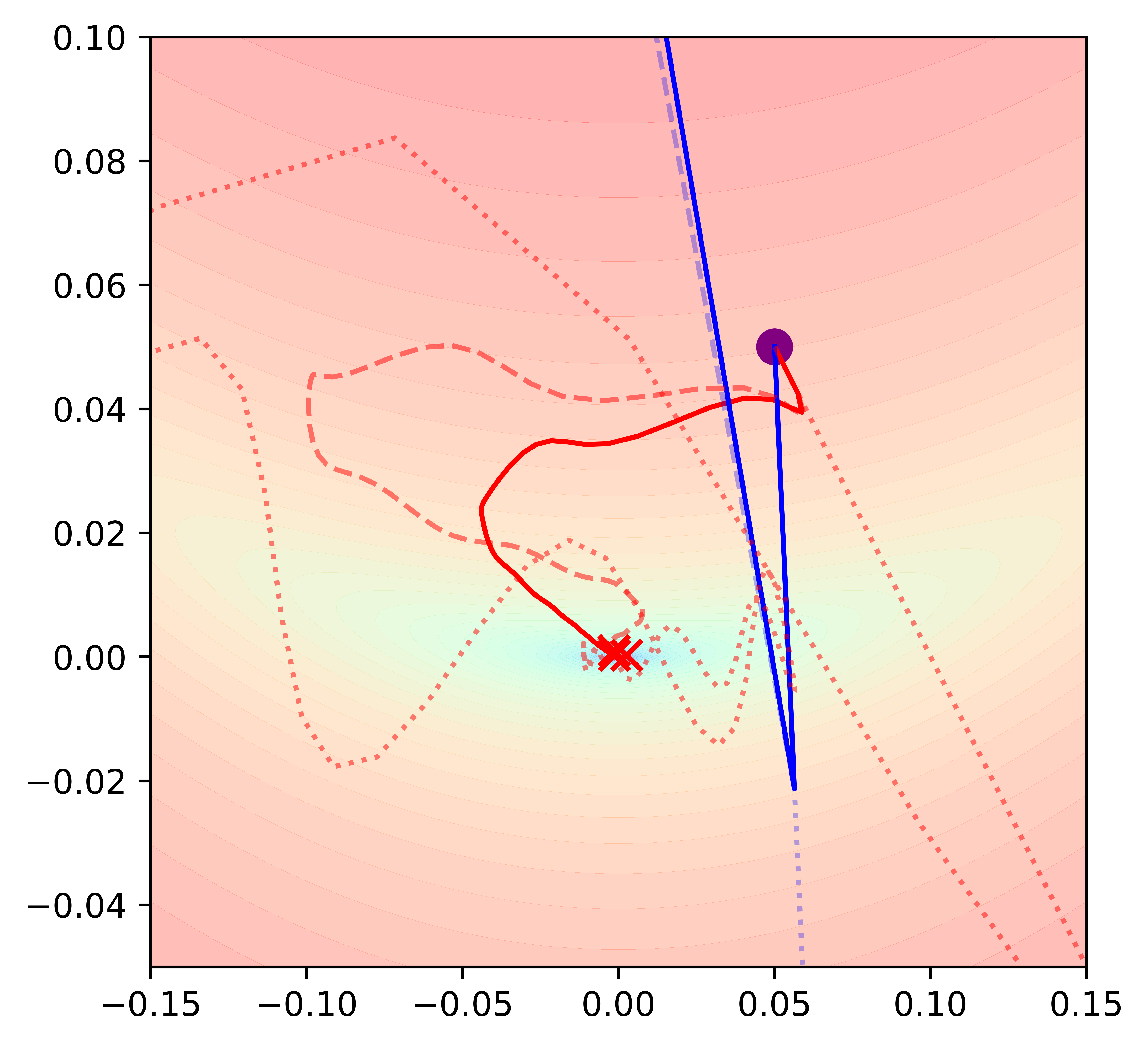} \\
    {\small lr = $0.15$}
    \end{tabular}
    \caption{Comparisons between \methodname{} or EMA + Adam, \methodname{} or EMA + SGD, Adam and SGD at different learning rates run on the Rosenbrock function. Parameter and function details are located in Appendix \ref{sec:lr_details}. }
    \label{fig:lr}
\end{figure}

\paragraph{Robustness to Learning Rate} First, we compare the performance of \methodname{} to EMA, Adam, and SGD on a set of ill-conditioned 2D examples. Figure \ref{fig:lr} shows how the stability of each optimizer varies w.r.t. the learning rate. As expected, we find that Adam is generally much more stable than SGD across different learning rates. In the high-learning rate regime, only \methodname{} and EMA, applied to Adam, are capable of converging without serious instability. In the medium learning rate regime, we see that weight averaging also reduces the tendency of the SGD update to diverge. Overall, weight averaging is better suited to stabilizing the trajectory of an optimizer than momentum alone, especially with higher learning rates.

This analysis is highly applicable to training deep neural networks due to the inherent link between learning rate, and loss function smoothness \cite{bottou2018optimization,wu2018wngrad}. Because the latter quantity may vary across weight values, it is important for an optimizer to be robust to learning rate, so that it is able to provide a stable minimizing trajectory across different levels of smoothness without sacrificing speed of convergence.

\begin{figure}[h!]
    \centering
    \begin{tabular}{cc}
        \includegraphics[width=.45\linewidth, trim={0 0 0 0}]{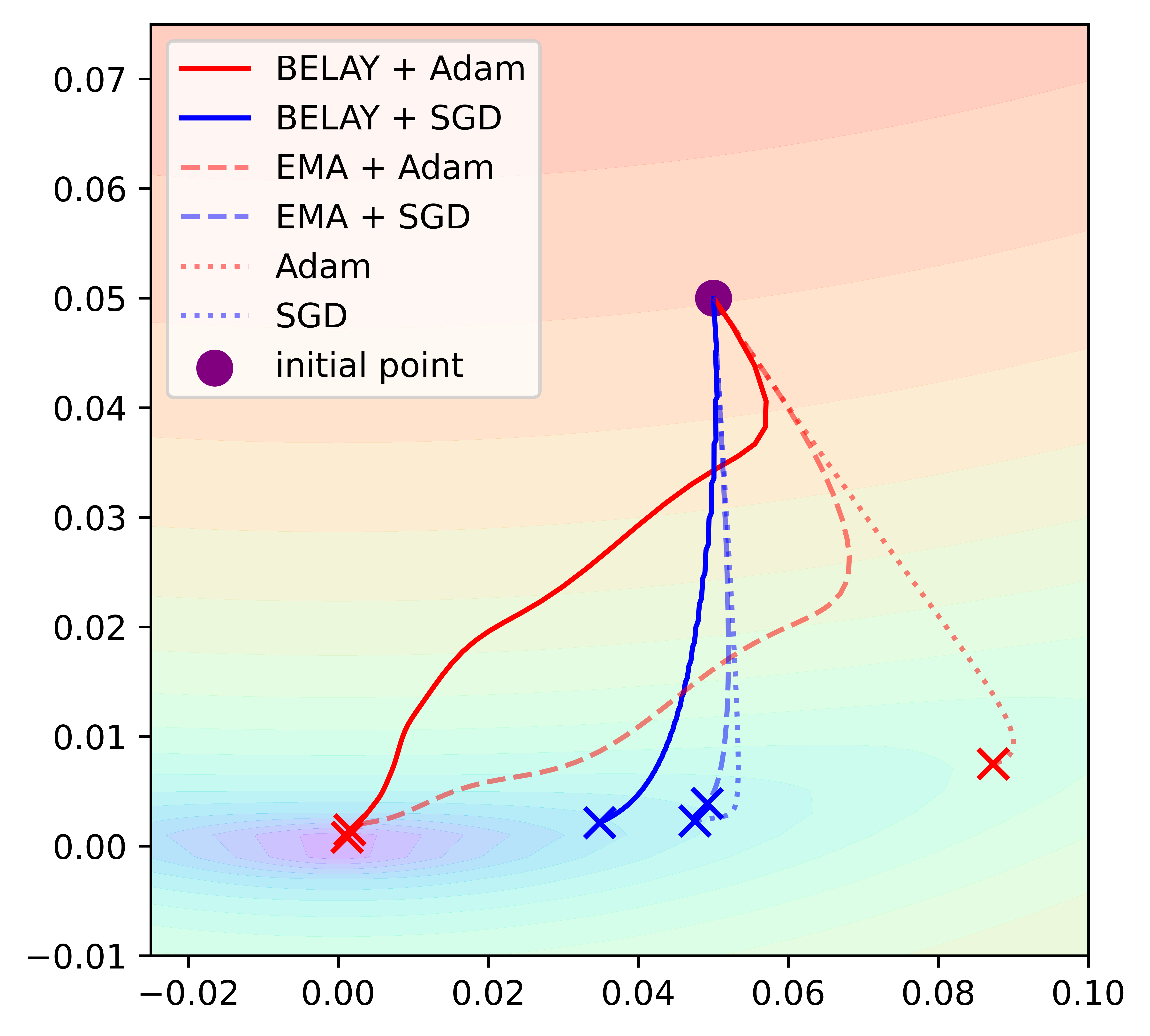} &    \includegraphics[width=.45\linewidth, trim={0 0 0 0}]{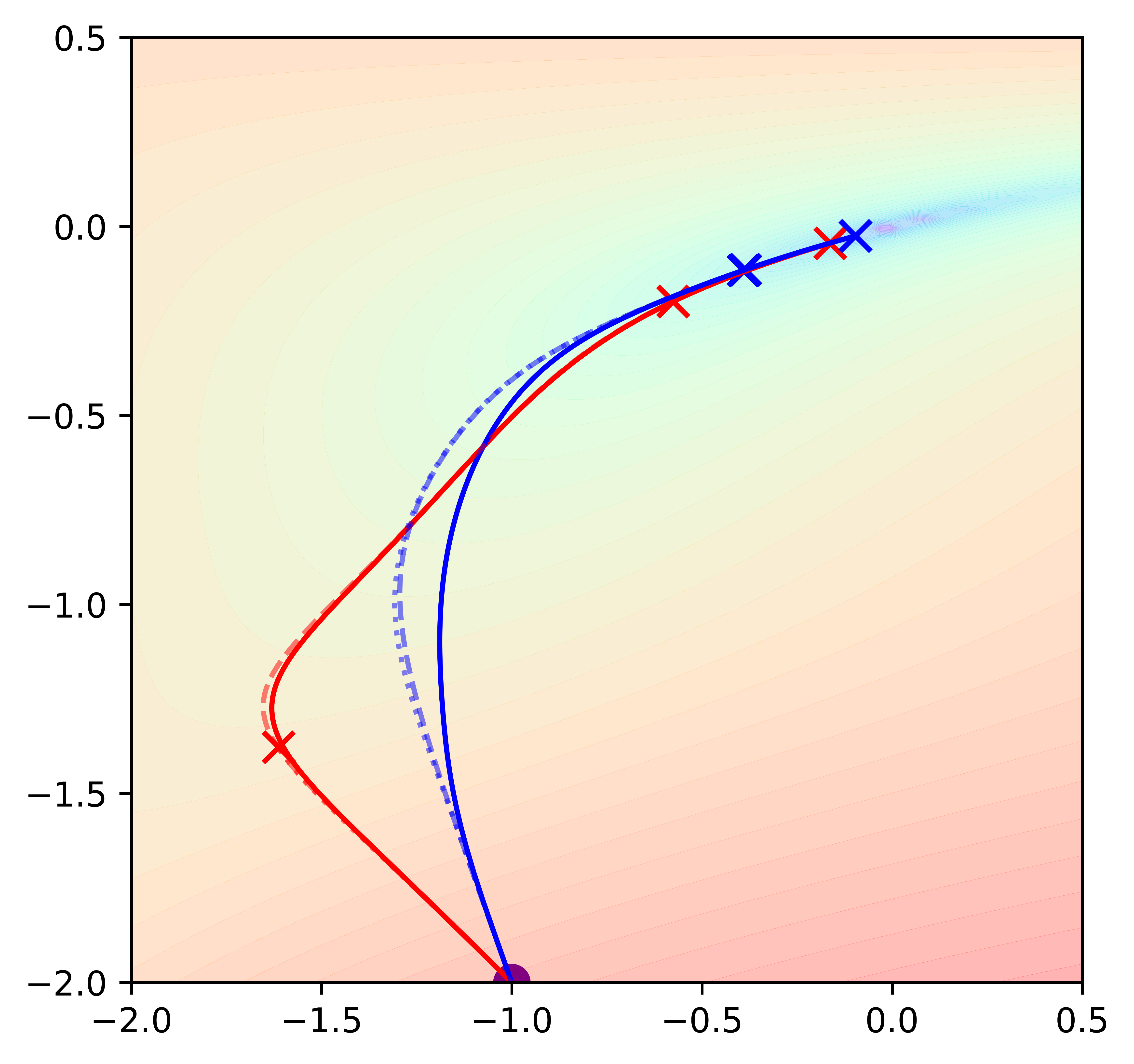}\\
        {\scriptsize Rosenbrock} & {\scriptsize Beale} \\
        \includegraphics[width=.45\linewidth, trim={0 0 0 0}]{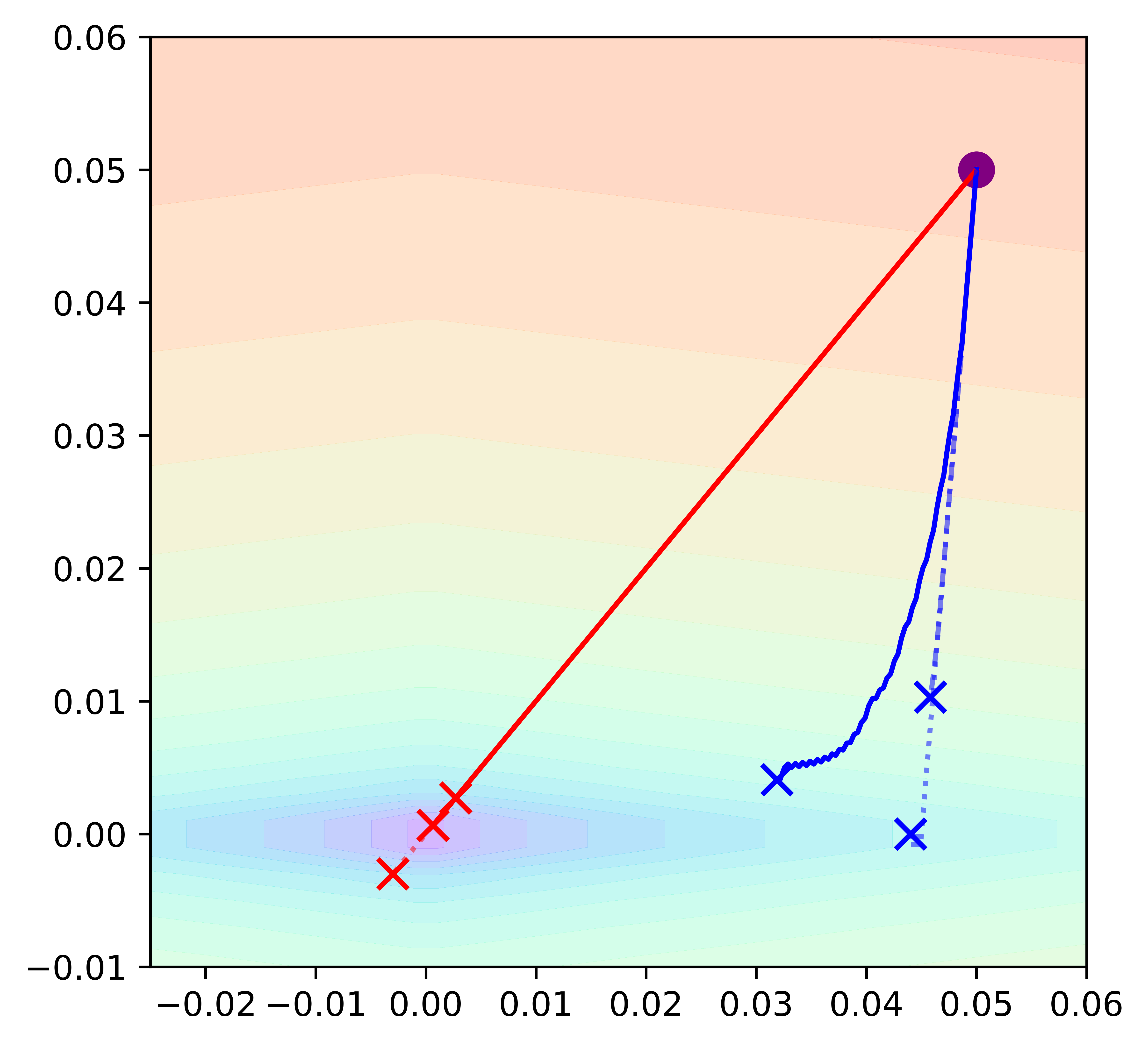} &   \includegraphics[width=.45\linewidth, trim={0 0 0 0}]{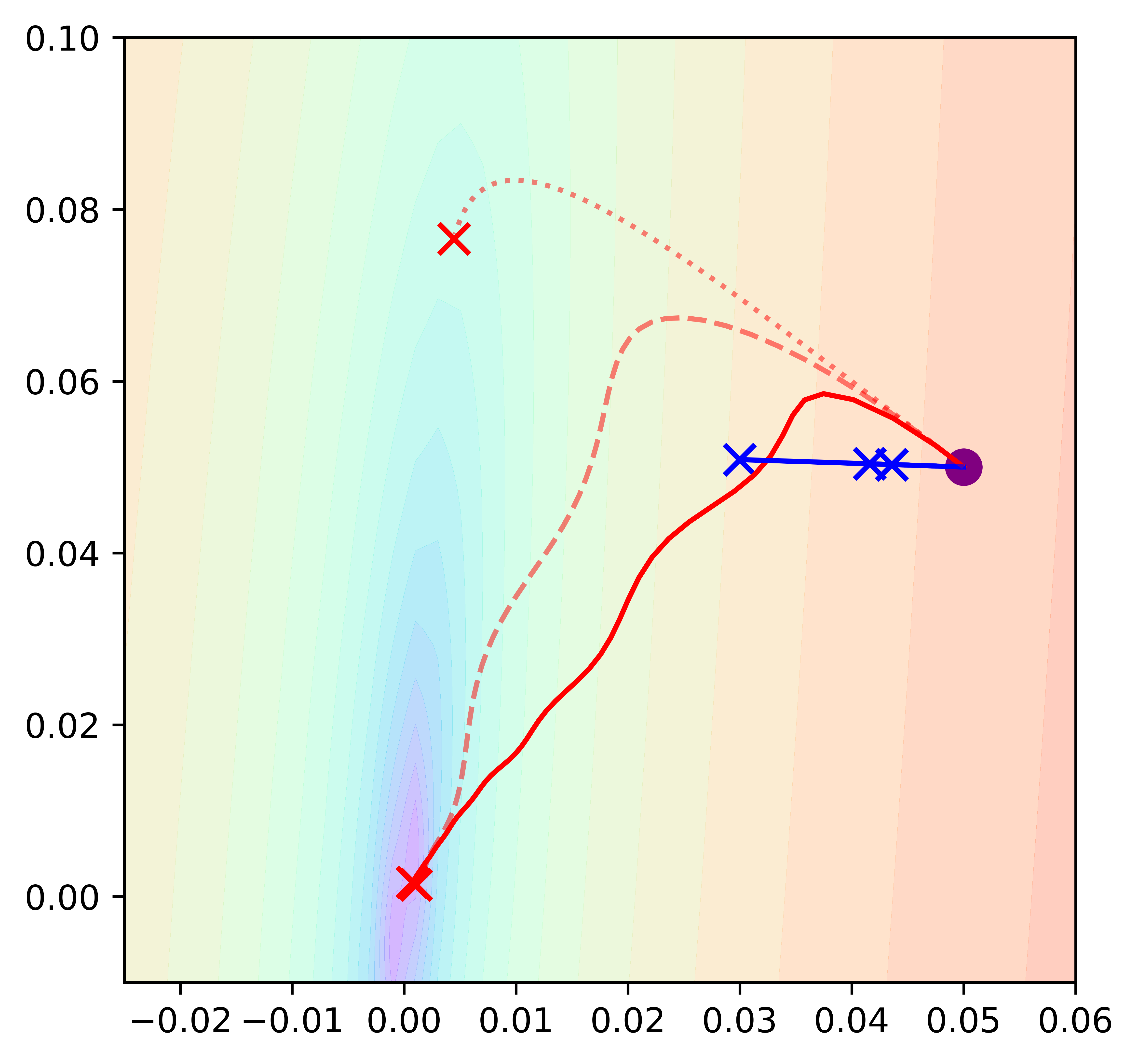} \\
        {\scriptsize $\frac{|x|}{10} + |y|$} & {\scriptsize $\frac{(x+\frac{y}{10})^2 + (x-\frac{y}{10})^2}{10}$}\\
    \end{tabular}
    \vspace{1em}
    \caption{Comparison of trajectories and speed of convergence for \methodname{}, EMA, and no weight averaging with both Adam and SGD. Each run is stopped near where \methodname{} has converged. Parameter and function details are located in Appendix \ref{sec:2d_details} }
    \label{fig:2dexp}
\end{figure}

\paragraph{Faster Convergence} We further analyze speed of convergence of the above algorithms in Figure \ref{fig:2dexp}. We find that \methodname{} with either Adam or SGD is able to consistently be among the fastest algorithms to converge. Note that in certain examples, such as the Beale function, momentum-based methods may struggle due to the existence of a nearby saddle-point (off to the left of the field of vision), however weight-averaging is able to remedy this.

\paragraph{Generative Modeling} Finally, we compare \methodname{} against EMA on a generative modeling task, on two datasets: CIFAR10 and MNIST. More experiments with generative modelling are located in Appendix  \ref{sec:cifar_details}. We see that \methodname{} compares favorably with respect to competing algorithms.

\begin{figure}[h!]
    \centering
    \begin{tabular}{c}
        \includegraphics[width=.9\linewidth, trim={0 0 0 0}, clip]{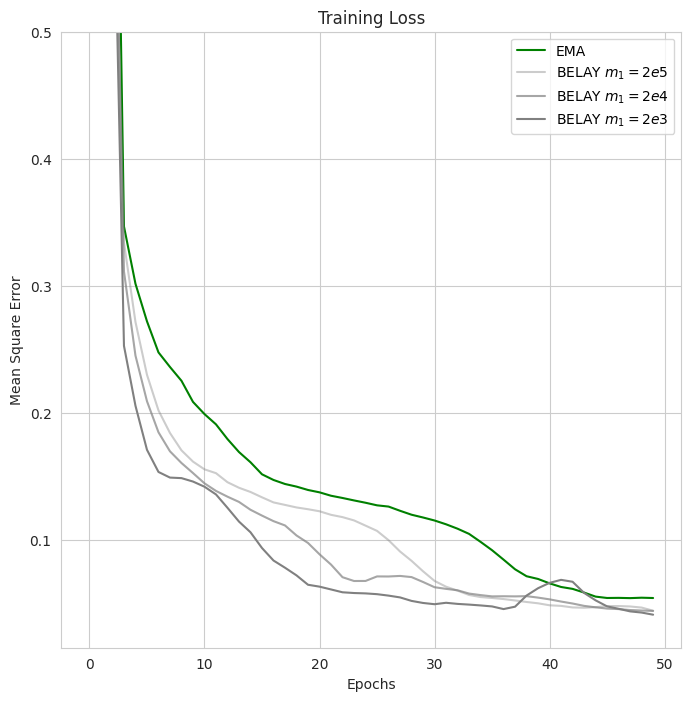}\\
        \includegraphics[width=.83\linewidth, trim={0.18cm 0 0 0}, clip]{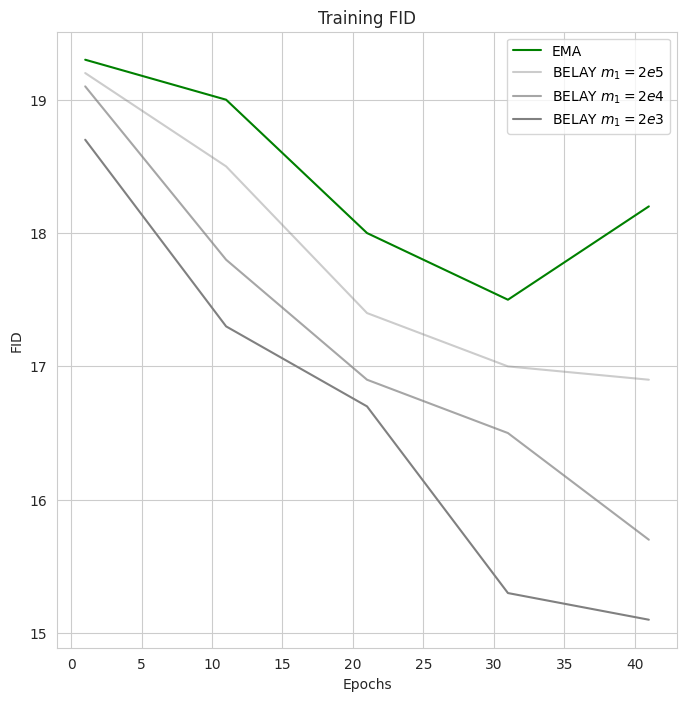}\\
    \end{tabular}
    \caption{Comparison between loss curves (top) and FID score across training iterations (bottom) of \methodname{} and EMA, trained to generate MNIST. Details in Appendix \ref{sec:mnist_details}}
    \label{fig:curves}
\end{figure}

\begin{table}[h]
    \centering
    \begin{tabular}{|c|c|c|c|}
        \hline
        \textbf{Method} & \textbf{Train Loss} & \textbf{Test Loss} & \textbf{FID}\\
        \hline
        \methodname{} & 0.042 & 0.040 & 15.2\\
        \hline
        EMA & 0.056 & 0.061 & 18.1 \\
        \hline
    \end{tabular}
    \caption{Comparison of final losses and FID scores of \methodname{} and EMA, run to generate MNIST digits. Details in Appendix \ref{sec:mnist_details}.}
    \label{MNIST-table}
\end{table}

\section{Conclusion and Future Work}
In this work, we have presented \methodname{}: a simple to implement and efficient weight-averaging method that uses physical systems to bridge the understanding between EMA and momentum based methods. We have shown how \methodname{} adds a new set of parameters to EMA, each with a strong and intuitive physical meaning, and in turn have shown how EMA uses momentum information, even when run on a non-momentum based method. Furthermore, we have shown empirically, that \methodname{} may outperform EMA in certain cases, implying that the choice of parameters is non-trivial. We have suggested some ways to set parameters to achieve desired behaviours.

With the basis outlined in this work, many future directions are apparent. First, by leveraging our newfound physical intuition, new theoretical guarantees for momentum-based methods as well as weight averaging could be drawn, by proving convergence bounds on \methodname{}. A deeper dive into the  damping parameter, and the interplay between the spring-based momentum term along with the classical heavy-ball momentum term of the loss function can be investigated to come up with what could be a more guaranteed "critically-damp" regime. Furthermore, large-scale empirical analysis could be performed to select optimal parameters for \methodname{}, and determine strong heuristics for practitioners interested in large-scale learning tasks.

\bibliography{main}

\begin{thebibliography}{28}
\providecommand{\natexlab}[1]{#1}
\providecommand{\url}[1]{\texttt{#1}}
\expandafter\ifx\csname urlstyle\endcsname\relax
  \providecommand{\doi}[1]{doi: #1}\else
  \providecommand{\doi}{doi: \begingroup \urlstyle{rm}\Url}\fi

\bibitem[Beck \& Teboulle(2003)Beck and Teboulle]{beck2003mirror}
Beck, A. and Teboulle, M.
\newblock Mirror descent and nonlinear projected subgradient methods for convex
  optimization.
\newblock \emph{Operations Research Letters}, 31\penalty0 (3):\penalty0
  167--175, 2003.

\bibitem[Bottou et~al.(2018)Bottou, Curtis, and
  Nocedal]{bottou2018optimization}
Bottou, L., Curtis, F.~E., and Nocedal, J.
\newblock Optimization methods for large-scale machine learning.
\newblock \emph{SIAM review}, 60\penalty0 (2):\penalty0 223--311, 2018.

\bibitem[Bussi \& Parrinello(2007)Bussi and Parrinello]{bussi2007accurate}
Bussi, G. and Parrinello, M.
\newblock Accurate sampling using langevin dynamics.
\newblock \emph{Physical Review E}, 75\penalty0 (5):\penalty0 056707, 2007.

\bibitem[Child et~al.(2019)Child, Gray, Radford, and
  Sutskever]{child2019generating}
Child, R., Gray, S., Radford, A., and Sutskever, I.
\newblock Generating long sequences with sparse transformers.
\newblock \emph{arXiv preprint arXiv:1904.10509}, 2019.

\bibitem[Dinh et~al.(2014)Dinh, Krueger, and Bengio]{dinh2014nice}
Dinh, L., Krueger, D., and Bengio, Y.
\newblock Nice: Non-linear independent components estimation.
\newblock \emph{arXiv preprint arXiv:1410.8516}, 2014.

\bibitem[Hestenes et~al.(1952)Hestenes, Stiefel, et~al.]{hestenes1952methods}
Hestenes, M.~R., Stiefel, E., et~al.
\newblock Methods of conjugate gradients for solving linear systems.
\newblock \emph{Journal of research of the National Bureau of Standards},
  49\penalty0 (6):\penalty0 409--436, 1952.

\bibitem[Hinton et~al.(2012)Hinton, Srivastava, and Swersky]{hinton2012neural}
Hinton, G., Srivastava, N., and Swersky, K.
\newblock Neural networks for machine learning lecture 6a overview of
  mini-batch gradient descent.
\newblock \emph{Cited on}, 14\penalty0 (8):\penalty0 2, 2012.

\bibitem[Ho et~al.(2019)Ho, Chen, Srinivas, Duan, and Abbeel]{ho2019flow++}
Ho, J., Chen, X., Srinivas, A., Duan, Y., and Abbeel, P.
\newblock Flow++: Improving flow-based generative models with variational
  dequantization and architecture design.
\newblock In \emph{International Conference on Machine Learning}, pp.\
  2722--2730. PMLR, 2019.

\bibitem[Ho et~al.(2020)Ho, Jain, and Abbeel]{ho2020denoising}
Ho, J., Jain, A., and Abbeel, P.
\newblock Denoising diffusion probabilistic models.
\newblock \emph{Advances in Neural Information Processing Systems},
  33:\penalty0 6840--6851, 2020.

\bibitem[Hyv{\"a}rinen \& Dayan(2005)Hyv{\"a}rinen and
  Dayan]{hyvarinen2005estimation}
Hyv{\"a}rinen, A. and Dayan, P.
\newblock Estimation of non-normalized statistical models by score matching.
\newblock \emph{Journal of Machine Learning Research}, 6\penalty0 (4), 2005.

\bibitem[Izmailov et~al.(2018)Izmailov, Podoprikhin, Garipov, Vetrov, and
  Wilson]{izmailov2018averaging}
Izmailov, P., Podoprikhin, D., Garipov, T., Vetrov, D., and Wilson, A.~G.
\newblock Averaging weights leads to wider optima and better generalization.
\newblock \emph{arXiv preprint arXiv:1803.05407}, 2018.

\bibitem[Karras et~al.(2022)Karras, Aittala, Aila, and
  Laine]{karras2022elucidating}
Karras, T., Aittala, M., Aila, T., and Laine, S.
\newblock Elucidating the design space of diffusion-based generative models.
\newblock \emph{arXiv preprint arXiv:2206.00364}, 2022.

\bibitem[Kingma \& Ba(2014)Kingma and Ba]{kingma2014adam}
Kingma, D.~P. and Ba, J.
\newblock Adam: A method for stochastic optimization.
\newblock \emph{arXiv preprint arXiv:1412.6980}, 2014.

\bibitem[Polyak(1964)]{polyak1964some}
Polyak, B.~T.
\newblock Some methods of speeding up the convergence of iteration methods.
\newblock \emph{Ussr computational mathematics and mathematical physics},
  4\penalty0 (5):\penalty0 1--17, 1964.

\bibitem[Polyak \& Juditsky(1992)Polyak and Juditsky]{polyak1992acceleration}
Polyak, B.~T. and Juditsky, A.~B.
\newblock Acceleration of stochastic approximation by averaging.
\newblock \emph{SIAM journal on control and optimization}, 30\penalty0
  (4):\penalty0 838--855, 1992.

\bibitem[Ruppert(1988)]{ruppert1988efficient}
Ruppert, D.
\newblock Efficient estimations from a slowly convergent robbins-monro process.
\newblock Technical report, Cornell University Operations Research and
  Industrial Engineering, 1988.

\bibitem[Salimans et~al.(2016)Salimans, Goodfellow, Zaremba, Cheung, Radford,
  and Chen]{salimans2016improved}
Salimans, T., Goodfellow, I., Zaremba, W., Cheung, V., Radford, A., and Chen,
  X.
\newblock Improved techniques for training gans.
\newblock \emph{Advances in neural information processing systems}, 29, 2016.

\bibitem[Salimans et~al.(2017)Salimans, Karpathy, Chen, and
  Kingma]{salimans2017pixelcnn++}
Salimans, T., Karpathy, A., Chen, X., and Kingma, D.~P.
\newblock Pixelcnn++: Improving the pixelcnn with discretized logistic mixture
  likelihood and other modifications.
\newblock \emph{arXiv preprint arXiv:1701.05517}, 2017.

\bibitem[Song \& Ermon(2020)Song and Ermon]{song2020improved}
Song, Y. and Ermon, S.
\newblock Improved techniques for training score-based generative models.
\newblock \emph{Advances in neural information processing systems},
  33:\penalty0 12438--12448, 2020.

\bibitem[Song et~al.(2020{\natexlab{a}})Song, Garg, Shi, and
  Ermon]{song2020sliced}
Song, Y., Garg, S., Shi, J., and Ermon, S.
\newblock Sliced score matching: A scalable approach to density and score
  estimation.
\newblock In \emph{Uncertainty in Artificial Intelligence}, pp.\  574--584.
  PMLR, 2020{\natexlab{a}}.

\bibitem[Song et~al.(2020{\natexlab{b}})Song, Sohl-Dickstein, Kingma, Kumar,
  Ermon, and Poole]{song2020score}
Song, Y., Sohl-Dickstein, J., Kingma, D.~P., Kumar, A., Ermon, S., and Poole,
  B.
\newblock Score-based generative modeling through stochastic differential
  equations.
\newblock \emph{arXiv preprint arXiv:2011.13456}, 2020{\natexlab{b}}.

\bibitem[Vincent(2011)]{vincent2011connection}
Vincent, P.
\newblock A connection between score matching and denoising autoencoders.
\newblock \emph{Neural computation}, 23\penalty0 (7):\penalty0 1661--1674,
  2011.

\bibitem[Welling \& Teh(2011)Welling and Teh]{welling2011bayesian}
Welling, M. and Teh, Y.~W.
\newblock Bayesian learning via stochastic gradient langevin dynamics.
\newblock In \emph{Proceedings of the 28th international conference on machine
  learning (ICML-11)}, pp.\  681--688, 2011.

\bibitem[Wu et~al.(2018)Wu, Ward, and Bottou]{wu2018wngrad}
Wu, X., Ward, R., and Bottou, L.
\newblock Wngrad: Learn the learning rate in gradient descent.
\newblock \emph{arXiv preprint arXiv:1803.02865}, 2018.

\bibitem[Yaz{\i}c{\i} et~al.(2018)Yaz{\i}c{\i}, Foo, Winkler, Yap, Piliouras,
  and Chandrasekhar]{yazici2018unusual}
Yaz{\i}c{\i}, Y., Foo, C.-S., Winkler, S., Yap, K.-H., Piliouras, G., and
  Chandrasekhar, V.
\newblock The unusual effectiveness of averaging in gan training.
\newblock \emph{arXiv preprint arXiv:1806.04498}, 2018.

\bibitem[Ypma(1995)]{ypma1995historical}
Ypma, T.~J.
\newblock Historical development of the newton--raphson method.
\newblock \emph{SIAM review}, 37\penalty0 (4):\penalty0 531--551, 1995.

\bibitem[Zeiler(2012)]{zeiler2012adadelta}
Zeiler, M.~D.
\newblock Adadelta: an adaptive learning rate method.
\newblock \emph{arXiv preprint arXiv:1212.5701}, 2012.

\bibitem[Zhuang et~al.(2020)Zhuang, Tang, Ding, Tatikonda, Dvornek,
  Papademetris, and Duncan]{zhuang2020adabelief}
Zhuang, J., Tang, T., Ding, Y., Tatikonda, S.~C., Dvornek, N., Papademetris,
  X., and Duncan, J.
\newblock Adabelief optimizer: Adapting stepsizes by the belief in observed
  gradients.
\newblock \emph{Advances in neural information processing systems},
  33:\penalty0 18795--18806, 2020.

\end{thebibliography}
\bibliographystyle{icml2023}

\onecolumn
\newpage
\twocolumn
\appendix

\section{Derivations}

\subsection{Derivation of $k$}
Letting $\bm{w}(0) = \bm{w}_0$ and $\dot{\bm{w}}_0 = 0$, we see that Eq. \ref{eq:overdamped_system} reduces to
\begin{align}
    \bm{w}_0 &= C_1 + C_2 \label{eq:x0} \\
    0 &= C_1\left(-\delta + \sqrt{\delta^2 - \frac{k}{m}} \right) + C_2\left(-\delta - \sqrt{\delta^2 - \frac{k}{m}} \right). \label{eq:v0}
\end{align}
Substituting Eq. \ref{eq:x0} into Eq. \ref{eq:v0}, we obtain
\begin{align*}
    C_1 = \bm{w}_0 \left( \frac{\delta}{\sqrt{\delta^2 - \frac{k}{m}}}  - \frac{1}{2} \right) \\
    C_2 = \bm{w}_0 \left(\frac{1}{2} - \frac{\delta}{\sqrt{\delta^2 - \frac{k}{m}}} \right).
\end{align*}

Now we see that the general solution of the harmonic oscillator takes the form
\begin{align*}
    \bm{w}(t) &= \bm{w}_0 \left( \frac{\delta}{\sqrt{\delta^2 - \frac{k}{m}}}  - \frac{1}{2} \right) e^{\left(-\delta + \sqrt{\delta^2 - \frac{k}{m}}\right) t}  \\ 
    &\hspace{.5in}+ \bm{w}_0 \left(\frac{1}{2} - \frac{\delta}{\sqrt{\delta^2 - \frac{k}{m}}} \right) e^{\left(-\delta - \sqrt{\delta^2 - \frac{k}{m}}\right) t}.
\end{align*}
Since all terms in our method are functions of $\Delta  t$, we may let $ \delta = \frac{1}{\Delta t} = 1$ without loss in generality. Letting $t = T$, we have
\begin{align*}
    \bm{w}(T) &= \bm{w}_0 \left( \frac{1}{\sqrt{1 - \frac{k}{m}}}  - \frac{1}{2} \right) e^{\left(-1 + \sqrt{1 - \frac{k}{m}}\right) T}  \\ 
    &\hspace{.5in}+ \bm{w}_0 \left(\frac{1}{2} - \frac{1}{\sqrt{1 - \frac{k}{m}}} \right) e^{\left(-1 - \sqrt{1 - \frac{k}{m}}\right) T}.
\end{align*}
We would like to choose $k = k(T)$ such that $\bm{w}(T) = \bm{w}(T_0)$
for some reference time $T_0$. We have found $T_0 = 1,000,000$ to be a reasonable default parameter. We see that we approximately satisfy this condition when $k = \frac{k_0}{T}$.

\section{Parameters}
We describe all the parameters of \methodname{} in detail with intuitive physical explanations, and appropriate units assigned in Table \ref{param-table}. We note that, outside of the optimizer $g(\bm{w}, t)$ and learning rate $\eta$, only $m_2$ is left as a free parameter in EMA. In \methodname{}, we additionally have $m_1 < \infty$ as a tunable parameter.

\begin{table}[h!]
    \centering
    \begin{tabular}{|p{1cm}|p{4cm}|p{1cm}|}
        \hline
        Param. & Description & SI Units\\
        \hline
        $k$ & spring constant, scales the force bringing both masses together. This force can also be controlled by changing both $m_1$ and $m_2$ simultaneously. & N/m\\
        \hline
        $m_1$ & mass of first point-mass, scales force on point-mass inversely. Implicitly built into $g'$, and scales momentum term if $c_1$ is not chosen to remove momentum. & g\\
        \hline
        $m_2$ & mass of second point-mass. Analogous to $m_1$ but on $\bm{w}_2$, but does not have an effect on $g'$ & g\\
        \hline
        $c_1, c_2$ & damping coefficients for both masses. Can be set to $\frac{2m}{\Delta t}$ in order to fully dampen the system and ensure a return to $0$ velocity between time-steps. Deviating will retain velocity and add a momentum term in an optimization sense.& Ns/m\\
        \hline
        $g(\bm{w},t)$ & weight update function, interpretable as a force applied to $\bm{w}_1$. This value is provided by an existing optimization algorithm. For example for full-batch gradient descent the value is $(-\nabla f)$ for some loss function $f$. & N\\
        \hline
        $\eta$ & scale of weight update, learning rate for gradient descent based methods. & $\text{s}^2$/m\\
        \hline
    \end{tabular}
    \caption{Physical parameters of \methodname{} and interpretations}
    \label{param-table}
\end{table}

\section{Implementation Details}

\subsection{Learning rate robustness experiment} \label{sec:lr_details}
We state all the parameter choices for all runs displayed in Figure \ref{fig:lr} here. Since we separately observe in Figure \ref{fig:lr} that different methods are robust to different learning rates, we run each optimizer at the highest learning rate that optimizer can handle while producing a stable converging trajectory across all runs.
\begin{enumerate}
    \item \methodname{} + Adam is run with the parameter set, $k=1$, $m_1=10$, $m_2=20$. Damping coeff. $c_1$, $c_2$ are set to zero out the velocity term as stated in Section \ref{sec:belay}
    \item \methodname{} + SGD is run with the parameter set, $k=1$, $m_1=10$, $m_2=20$. Damping coeff. $c_1$, $c_2$ are set to zero out the velocity term as stated in Section \ref{sec:belay}
    \item EMA + Adam is run with the parameter $\alpha=0.95$.
    \item EMA + SGD is run with the parameter $\alpha=0.95$.
    \item Adam is run with default parameters.
\end{enumerate}

The Rosenbrock function was chosen with parameters $a=0$, $b=100$ for easy visualization:
\[
    f(x,y) = (a - x)^2 + b(x - (y^2))^2.
\]

The following version of the Beale function was used, centering the global minima at (0,0) for easy visualization:

\begin{align*}
    f(x,y) &= (1.5 - (x+3)+ (x+3)*(y+0.5))^2\\
    &\quad + (2.25 - (x+3) + (x+3)*(y+0.5)^2)^2\\
    &\quad + (2.625 - (x+3) + (x+3)*(y+0.5)^3)^2
\end{align*}

\subsection{Convergence trajectory experiments} \label{sec:2d_details}
We state all the learning rate choices for all runs displayed in Figure \ref{fig:2dexp} here. The values for all parameters except for learning rate were the same as those described in Appendix \ref{sec:lr_details}.
\begin{enumerate}
    \item \methodname{} + Adam is run with learning rate $\eta = 5*10^{-2}$.
    \item \methodname{} + SGD is run with $\eta = 10^{-2}$.
    \item EMA + Adam is run with $\eta = 10^{-2}$.
    \item EMA + SGD is run with $\eta = 10^{-3}$.
    \item Adam is run with $\eta = 10^{-3}$
    \item SGD is run with $\eta = 10^{-3}$
\end{enumerate}

\subsection{MNIST Experiment} \label{sec:mnist_details}

\methodname{} is run with parameters $k=1$, $m_1=2000$, $m_2=500$, and damping $c_1$, $c_2$, to zero out the velocity term as described in Section \ref{sec:belay}. The Adam optimizer is used with learning rate $\eta = 5*10^{-2}$. Both methods are run to optimize the score-based diffusion model described in \cite{song2020improved} on MNIST digits.


\begin{figure}[h]
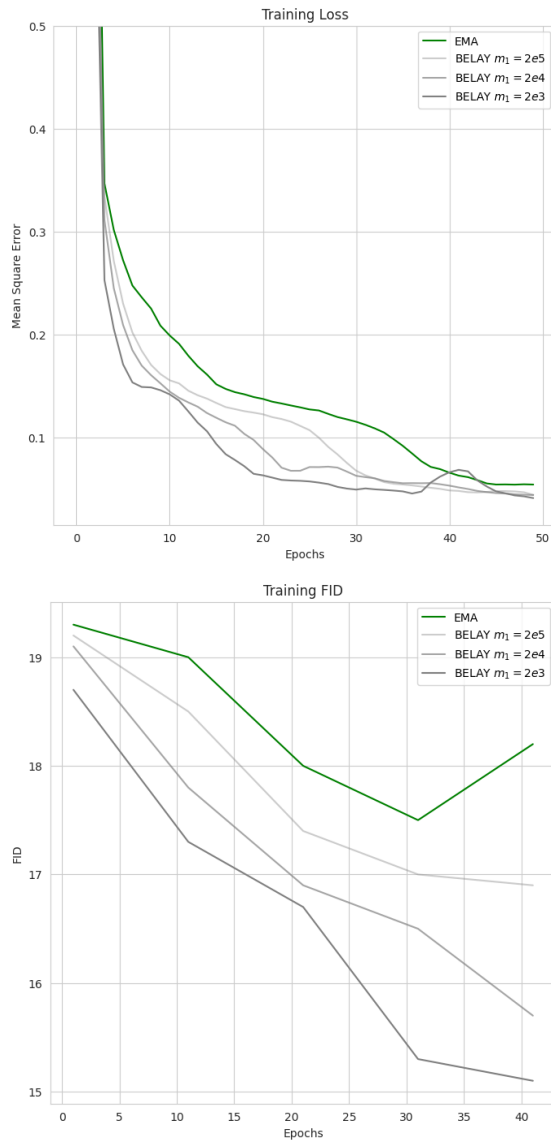

    \centering
    \begin{tabular}{c}
        \includegraphics[width = .9\linewidth]{figures_data/ema_loss_upres.png}\\
        \includegraphics[width = .9\linewidth]{figures_data/ema_fid_upres.png}\\
    \end{tabular}
    \caption{Comparison of train (top) and test loss (bottom) across training iteration of \methodname{} at parameters $m_1=2000, 2*10^{4}, 2*10^{5}$ and EMA, trained to generate MNIST with the score-based diffusion model from \cite{song2020improved}}
    \label{fig:mnist_apndx}
\end{figure}

\subsection{CIFAR Experiment} \label{sec:cifar_details}

\begin{table}[h]
    \centering
    \begin{tabular}{|c|c|c|c|}
        \hline
        \textbf{Method} & \textbf{Train Loss} & \textbf{Test Loss} & \textbf{FID}\\
        \hline
        \methodname{} & 0.176 & 0.170 & 1.99 \\
        \hline
        EMA & 0.175 & 0.168 & 1.98 \\
        \hline
    \end{tabular}
    \caption{Comparison of final losses and FID scores of \methodname{} and EMA, run to generate CIFAR-10 images.}
\end{table}

\methodname{} is run with parameters $(k=1, m_1=20000, m_2=2000)$, and damping $c_1$, $c_2$, to zero out the velocity term as described in Section \ref{sec:belay}. Both methods are run to optimize the score-based diffusion model described in \cite{karras2022elucidating} on CIFAR-10 images.


\onecolumn
\begin{figure}
    \centering
    \includegraphics[width=0.9\textwidth]{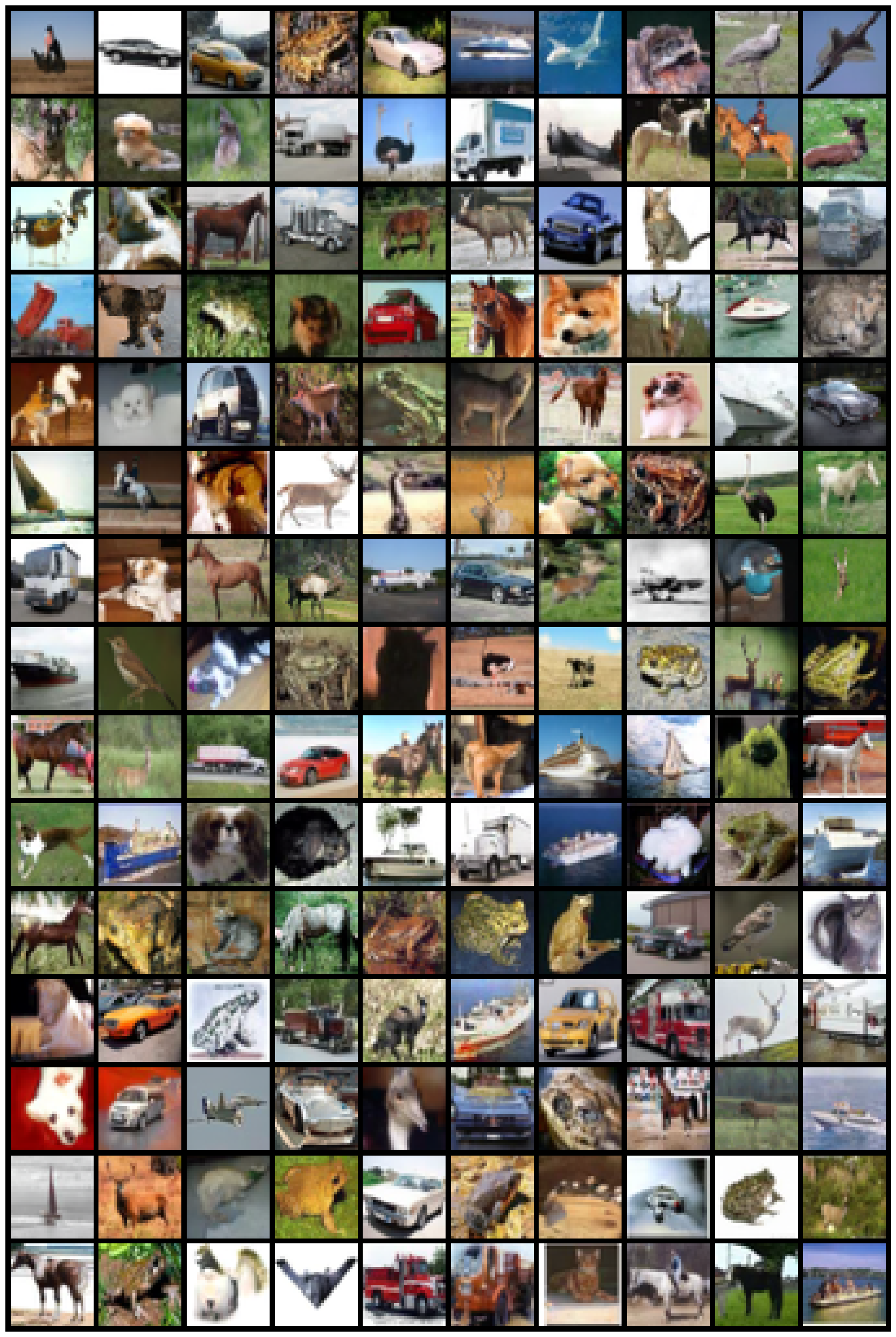}
    \caption{Images from an unconditional diffusion model with EMA trained on the CIFAR10 dataset.}
\end{figure}

\begin{figure}
    \centering
    \includegraphics[width=0.9\textwidth]{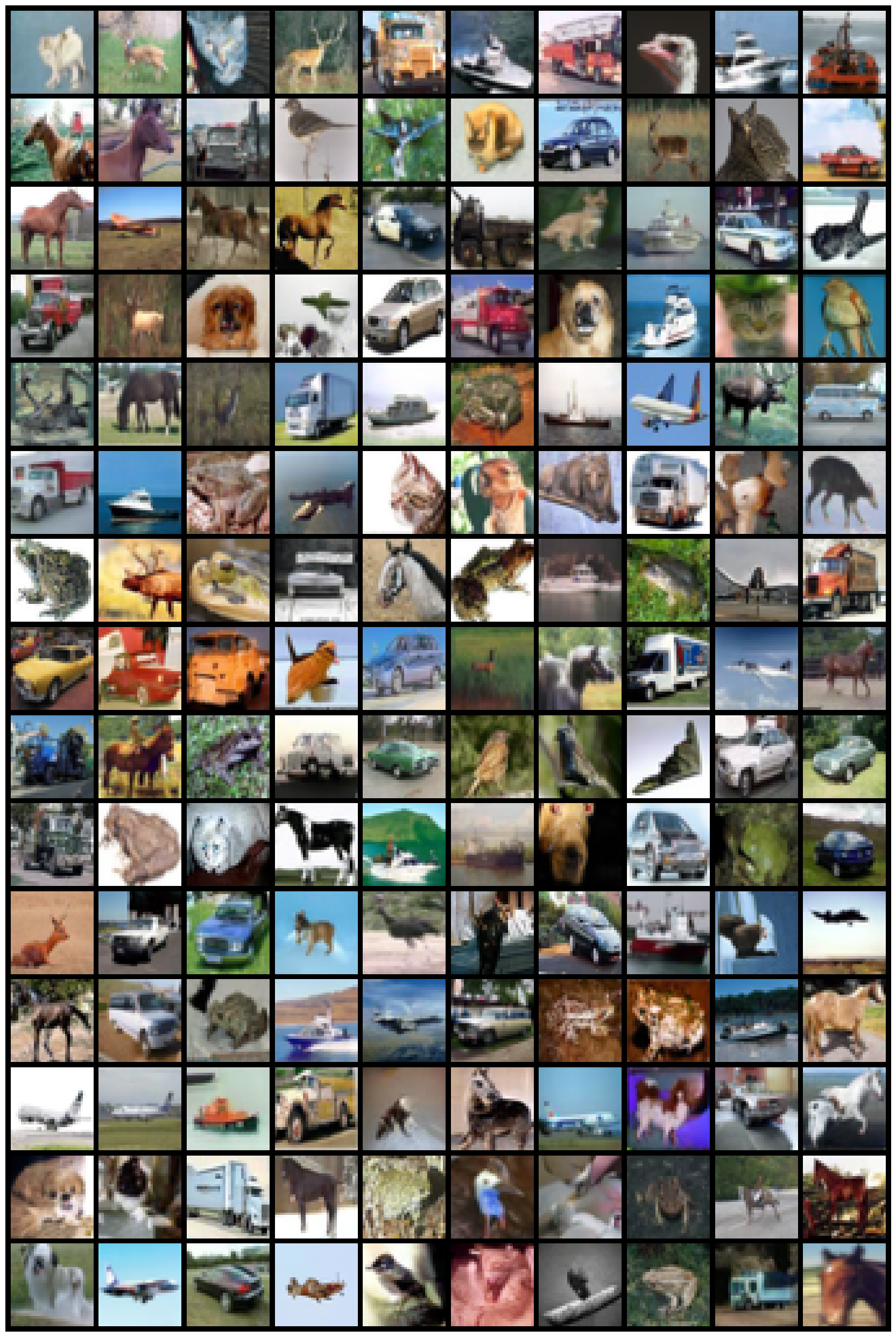}
    \caption{Images from an unconditional diffusion model with \methodname{} trained on the CIFAR10 dataset.}
\end{figure}


\end{document}